\documentclass{article}
\usepackage{spconf,amsmath,graphicx,enumitem,bm,balance}
\usepackage[htt]{hyphenat}
\usepackage{subcaption}

\title{Investigating context features hidden in End-to-End TTS}
\name{Kohki Mametani, Tsuneo Kato, Seiichi Yamamoto\thanks{This work was supported by JSPS KAKENHI Grant Number
17K02954.}}
\address{Department of Intelligent Information Engineering and Sciences, Doshisha University, Kyoto, Japan}
\begin{document}
\ninept
\maketitle
\begin{abstract}
Recent studies have introduced end-to-end TTS, which integrates the production of context and acoustic features in statistical parametric speech synthesis. As a result, a single neural network replaced laborious feature engineering with automated feature learning. However, little is known about what types of context information end-to-end TTS extracts from text input before synthesizing speech, and the previous knowledge about context features is barely utilized.
In this work, we first point out the model similarity between end-to-end TTS and parametric TTS. Based on the similarity, we evaluate the quality of encoder outputs from an end-to-end TTS system against eight criteria that are derived from a standard set of context information used in parametric TTS. We conduct experiments using an evaluation procedure that has been newly developed in the machine learning literature for quantitative analysis of neural representations, while adapting it to the TTS domain. Experimental results show that the encoder outputs reflect both linguistic and phonetic contexts, such as vowel reduction at phoneme level, lexical stress at syllable level, and part-of-speech at word level, possibly due to the joint optimization of context and acoustic features.
\end{abstract}
\section{Introduction}
\label{sec:intro}
Statistical parametric speech synthesis~\cite{spss2009} has steadily advanced through the history of annual Blizzard Challenges~\cite{king2014}, and the Hidden Markov Model (HMM)-based speech synthesis system (HTS)~\cite{hts} has been a dominant framework in this approach. Since the first release of the HTS, acoustic modeling in this approach has markedly improved due to the progress of its generative model from HMM to deep neural network~\cite{zen2013dnn, zen2014dnn} and recurrent neural network, especially long short-term memory (LSTM)~\cite{fan2014lstm, zen2015lstm}.  In contrast, little progress can be seen in text analysis, or "front end." Due to the very weak connection between text and speech, the front end extracts context features (also known as linguistic features) which are useful to bridge the gap between the two modalities. Conventionally, a standard set of context features gives a wide range of context information within a given text to an acoustic model, extensively covering phonetic, linguistic, and prosodic contexts~\cite{htslabel2015}.

Beyond the partial use of neural networks for an acoustic model, recent studies have introduced fully neural TTS systems, known as end-to-end TTS systems, which can be trained in an end-to-end fashion, requiring only pairs of an utterance and its transcript. These systems have already outperformed parametric TTS systems in terms of naturalness~\cite{tacotron}. In addition, acoustic modeling and text processing in a parametric TTS system are integrated by a single neural network. As a result, this singular solution expels language-specific knowledge used for the configuration of text analysis and speech-specific tasks to build an acoustic model, such as segmenting and aligning audio files, making it significantly easier to develop a new TTS system. 
Additionally, this allows such models to be conditioned on various attributes such as the speaker's prosodic feature~\cite{yuxuan2018}, enabling a truly joint optimization over context and acoustic features. As shown by the opening of the Blizzard Machine Learning Challenge~\cite{bmlc2017}, TTS has partly become a subject of machine learning and is expected to move on to the end-to-end style.

These advantages, however, often come at the cost of model interpretability. 
Understanding of the internal process of end-to-end TTS systems is difficult because neural networks are generally {\it black boxes}, making the functionality of systems based on such models unexplainable to humans. Therefore, model interpretability is essential to establish a more informed research process and improve current systems. In this vein, recently, a unified procedure for quantitative analysis of internal representations in end-to-end neural models has been developed. 
In~\cite{belinkov-asr}, hidden representations in an end-to-end automatic speech recognition system are thoroughly analyzed with the method, and it reveals the extent to which a character-based connectionist temporal classification model uses phonemes as an internal representation. Also, in~\cite{belinkov-nmt}, the same evaluation process is applied to analyze internal representations from different layers of a neural machine translation model. 

In this work, by adapting this evaluation procedure to the TTS domain, we demonstrate what types of context information are utilized in end-to-end TTS systems. We meta-analytically sort out the eight most important context features from the standard feature set in parametric TTS and use them as criteria for our experiments to quantify how and to what extent encoder outputs correlate with such context features. Specifically, unlike speech recognition and machine translation tasks, the performance of TTS systems has been primarily evaluated using subject tests such as Mean Opinion Score which often takes a lot of time and resources. For this reason, there are benefits to exploring a more convenient and objective evaluation process and investigating its usefulness for the further success of end-to-end TTS research. 

\section{Model Similarity}
In spite of the difference of the generative model in use, the way end-to-end TTS synthesizes speech is comparable to the way parametric TTS does as both approaches are categorized into the generative model-based TTS~\cite{zen2017video}. In the following explanation, we formally describe text input as $\bm{w} = \{\bm{w_i}\ |\ i = 1,2,...,L\}$ and time-domain speech output as $\bm{x} = \{\bm{x_j}\ |\ j = 1, 2,...,T\}$, where $L$ is the length of symbols in the text and $T$ is the number of frames of the speech waveform.

\begin{figure}[t!]
	\begin{center}
		\includegraphics[width=85mm]{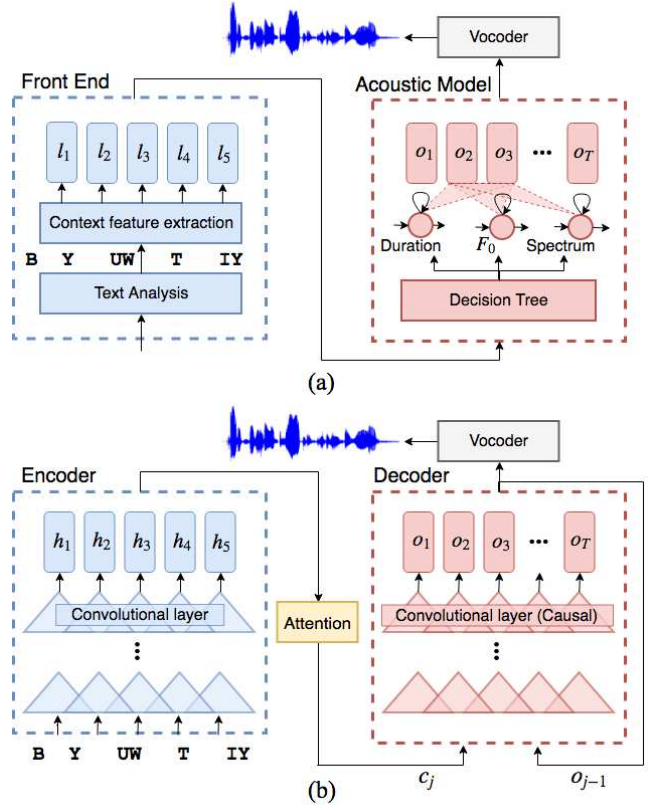}
	\end{center}
	\caption{Overview of speech synthesis process of (a) parametric TTS consisting of front end, acoustic model illustrated as HMMs, and vocoder and (b) end-to-end TTS model consisting of encoder, attention-based decoder illustrated as causal convolution networks, and vocoder.}
	\label{schemes}
\end{figure}

Fig.~\ref{schemes} (a) shows a typical speech synthesis process of the HTS, which represents parametric TTS, and it can be mainly divided into three steps. First, a front end extracts linguistic and phonetic contexts as well as prosodic ones at each of the phonemes within the text $\bm{w}$ and accordingly assigns context features $\bm{l} = \{\bm{l_i}\ |\ i = 1,2,...,L\}$. Typically, a context feature $\bm{l_i}$ is composed of a high dimensional vector, e.g., a 687-dimensional vector is used in the HTS-2.3.1. 
Second, an acoustic model generates acoustic features $\bm{o} = \{\bm{o_j}\ |\ j = 1,2,...,T\}$ for given context features $\bm{l}$, estimating features such as spectrum, $F_0$, and duration with individually clustered context-dependent HMMs. 
%As the HTS employs these context features to construct a number of questions for the decision tree within its HMMs, therefore, the number of such features needs to be kept as small as possible so that the following HMMs can be trained effectively.
Lastly, a vocoder synthesizes a real-time waveform $\bm{x}$ from acoustic features $\bm{o}$.

Fig.~\ref{schemes} (b) illustrates an end-to-end TTS model that achieves the integration of the production of context and acoustic features with a nonlinear function. Most of the end-to-end TTS models utilize an attention-based encoder-decoder framework~\cite{bahdanau2014}, directly mapping text to a speech waveform. The implementation of the framework varies depending on the generative model in its decoder, such as LSTM~\cite{tacotron,tacotron2} or causal convolution~\cite{tachibana2017, deepvoice3}, modeling temporal dynamic behaviors of speech.
First, an encoder encodes the text $\bm{w}$, folding context information around each symbol $\bm{w_i}$ and turning it into the corresponding high dimensional vector representations $\bm{h} = \{\bm{h_i}\ |\ i=1,2,...,L\}$.
Then, this is followed by a decoder with an attention mechanism. Before decoding the encoder outputs, all $\bm{h}$ is fed to an attention mechanism. At each decoding step $j$, the attention produces a single vector $\bm{c_j}$, known as the context vector, by computing the weighted sum of the sequence of the encoder outputs $\bm{h_i}$ (called alignments) as follows, where $\alpha_{ij}$ is a real value [0, 1]:

\[
	\bm{c_j} = \Sigma_{i=1}^{L} \alpha_{ij} \bm{h_i}
\]

The context vector summarizes the most important part of the encoder outputs for the current decoding step $j$. Although the computation of alignments varies depending on the system, the differences are trivial in the scheme. Then, the decoder takes the context vector $\bm{c_j}$ and the previous decoder output $\bm{o_{j-1}}$ as input and generates an acoustic feature $\bm{o_j}$, and finally a vocoder is used in the same way as in parametric TTS.
 
By using the one-to-one model comparison between the two models above, it is shown that both $\bm{l}$ in parametric TTS and $\bm{h}$ in end-to-end TTS play a similar role: converting each symbol in the text $\bm{w}$ into a high dimensional vector within the corresponding model. Both of them represent context information given to each symbol. This is the focus of our work. Our hypothesis is that the encoder outputs $\bm{h}$ contain the same type of context information utilized in $\bm{l}$ of parametric TTS. Moreover, due to the joint optimization with acoustic features, $\bm{h}$ should embrace extra details that are not seen in $\bm{l}$ such as ones caused by articulation, allowing it to be more effective context features for the overall  performance.

\section{Context features in parametric TTS}
\label{sec:lfeat}
In statistical parametric speech synthesis, there are several studies that investigate the quality of the standard set of context features. The contribution of higher-level context features, such as part of speech and intonational phrase boundaries, has been studied~\cite{WattsIS2010}. This study reveals that features above word level have no significant impact on the quality of synthesized speech. In \cite{hengluIS2012}, a Bayesian network is used to evaluate how each of the 26 commonly used context features in the standard set contributes to several aspects of acoustic features. This revealed the most important context features that are relevant to three acoustic features (i.e., spectrum, $F_0$, and duration), the features relevant to the acoustic features except for spectrum, and the features relevant to either $F_0$ or duration. By applying a smaller feature set while removing irrelevant context features, it is demonstrated that a parametric TTS system with fewer contexts can produce a speech waveform with a quality that is as good as that of the contextually rich system. 
As for the representation of positional features, \cite{rasmus2016} explores the advantages of categorical and relative representations against the absolute representation used in standard models. In the study, four categories are proposed to represent positional values: "beginning" for the first element, "end" for the last element, "one" for the segments of length one, and "middle" for all the others. It turns out that a system with categorical representation generates the best speech quality among other representations. 

Originally, 11 features are confirmed to be important~\cite{hengluIS2012}. The set of features includes two pairs of positional features, which only differ by whether it counts forward or backward. The difference can be canceled by using the aforementioned categorical representation, resulting in a reduction of two features. In addition, the accent is considered synonymous with stress in our work. As a result, the remaining eight features can be summarized (Table~\ref{table:contexts}).

\section{Experiments}
\subsection{Methodology}
\begin{figure}[t!]
	\begin{center}
		\includegraphics[width=85mm]{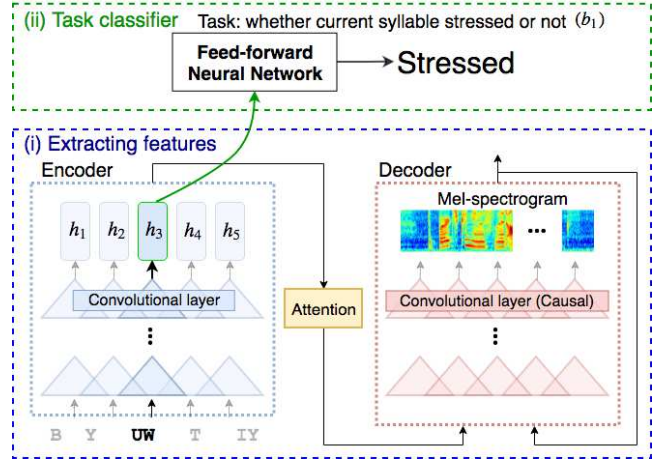}
	\end{center}
	\caption{Illustration of our evaluation process. After training encoder and decoder of end-to-end TTS, we (i) extract encoder outputs (e.g., $h_3$) and (ii) train supervised classifier on certain task using extracted representations and evaluate its performance.}
	\label{proposal}
\end{figure}

Recently, a unified procedure for quantitative analysis of internal representations in end-to-end neural models has been developed~\cite{belinkov2018}. In our work, we apply this procedure to analyze the feature representations learned by an encoder in end-to-end TTS.
Fig.~\ref{proposal} shows our evaluation process. After training an end-to-end model, we save its learned parameters and create a pre-trained model. Then, we dynamically extract the values from the computational graph of the model in order to collect a number of its encoder outputs. With the extracted representations, we follow a basic process of multi-class classification task: training a classifier on a simple supervised task using the encoder outputs and then evaluating the performance of the classifier.
We assume that if a feature related to the classification task is hidden in the encoder outputs, it will work as evidence for classification, and the classifier's performance will be increased. In this manner, the performance of the trained classifier can be used as a proxy for an objective quality of the representations.
Since the procedure assesses only one aspect of such representations per classification task, the choice of criterion with which the classifier classifies its input needs careful consideration. In this preliminary work,
we start with the eight contexts in Table~\ref{table:contexts} as evaluation criteria of the classification and iterate the experiment eight times while changing the criterion and accordingly adjusting the size of the classifier's output.

\begin{table}[]
\begin{tabular}{|c|c|c|}\hline
ID            & Context information                         & Card.\\\hline
$p_2$         & previous phoneme identity                   & 39\\\hline
$p_3$         & current phoneme identity                    & 39\\\hline
$p_4$         & next phoneme identity                       & 39\\\hline
$p_6$ (=$p_7$)& position of current phoneme in syllable     & 4\\\hline
$b_1$ (=$b_2$)& whether current syllable stressed or not    & 2\\\hline
$b_4$ (=$b_5$)& position of current syllable in word        & 4\\\hline
$b_{16}$      & name of vowel of current syllable           & 15\\\hline
$e_1$         & gpos (guess part-of-speech) of current word & 8\\\hline
\end{tabular}
\caption{Essential context features for parametric TTS and their cardinalities (Card.). The same ID is given to each feature as in~\cite{hengluIS2012}}
\label{table:contexts}
\end{table}

\subsection{Experimental Setup}
The end-to-end TTS model used in our experiments is a well-known open source PyTorch implementation\footnote{Audio samples are available: https://r9y9.github.io/deepvoice3\_pytorch/} of Baidu's Deep Voice 3~\cite{deepvoice3}. The model is trained on the LJ Speech Dataset~\cite{ljspeech17}, a public domain speech dataset consisting of 13,100 pairs of a short English speech and its transcript. To make the input format correspond to parametric TTS, we build a model that takes only phonemes as input by simply converting the words in the transcripts to their phonetic representations (ARPABET) during a preprocessing step.
After training the model, we synthesize speech based on 25,000 short US English sentences from the M-AILABS Speech Dataset~\cite{mailabs} while collecting its encoded phoneme representations (encoder outputs).
Depending on the classifier's criterion, each encoder output is assigned a correct label. Lexical stress is given by looking up the word in the CMU Pronouncing Dictionary, syllabication of each word is performed using an open-source tool~\footnote{https://github.com/kylebgorman/syllabify}, and part of speech tags are assigned by a pre-trained POS tagger developed in the Penn Treebank project using eight coarse-grained fundamental tags (excluding "interjection" because of its scarcity).
Then, the encoder outputs are split into training and test sets for the classifier in the ratio of 80/20, and finally, we evaluate the classification performance to obtain a quantitative measure of the feature representations about the given contextual criterion.
The implementation of the classifier is made to be as simple as the one suggested in previous studies~\cite{belinkov-asr, belinkov-nmt}. The size of the input to the classifier is 128, which is equal to the dimension of the encoder output of the TTS model. Our classifier is a feed-forward neural network with one hidden layer, where the size of the hidden layer is set to 64. This is followed by a ReLU non-linear activation and then a softmax layer mapping onto the label set size, which is dependent on the cardinality of the context.

\begin{figure*}[t!]
	\begin{center}
		\includegraphics[width=180mm]{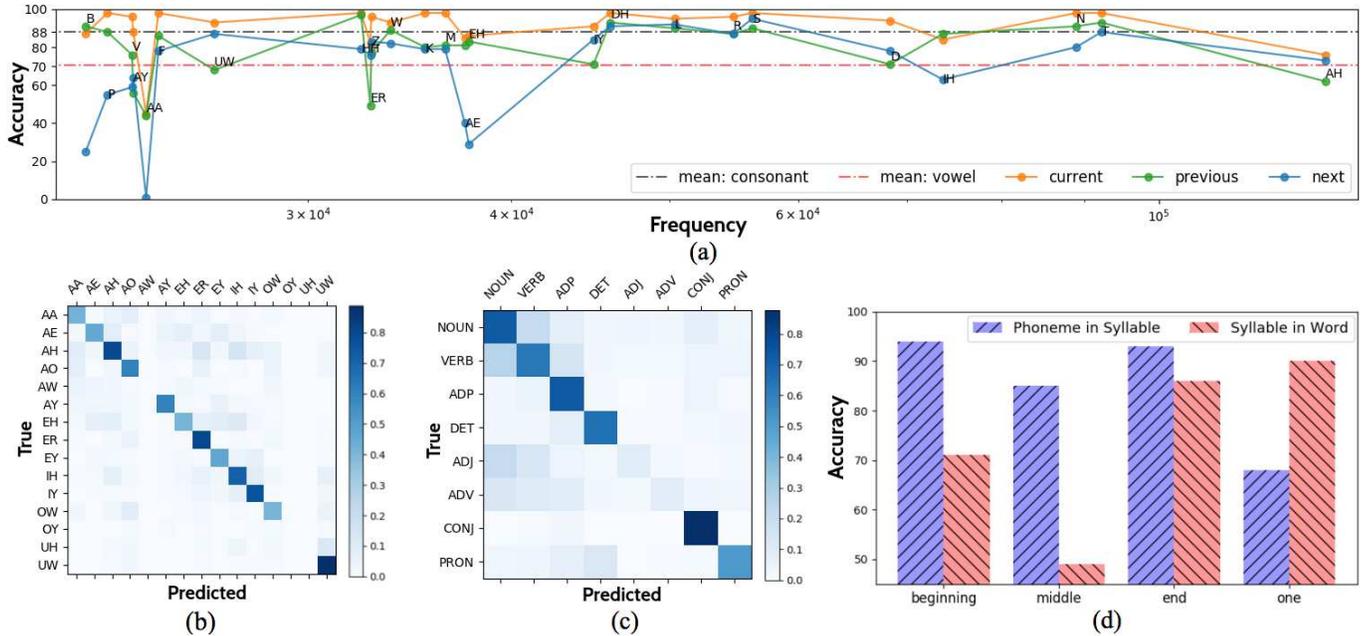}
	\end{center}
	\caption{{\it \textbf{Upper}}: (a) Classification accuracy of phonemes per appearance frequency. Phonemes that make up $90\%$ of total phoneme appearance in training data are displayed. {\it \textbf{From bottom left to right}}: (b) confusion matrices for name of vowel of current syllable and (c) for POS tagging and (d) comparison of prediction accuracy of positional features at level of syllable and word.}
	\label{bigresults}
\end{figure*}

\subsection{Results}
\subsubsection{Evaluation of phoneme identities ($p_2$, $p_3$, $p_4$)}
Phoneme identity is the most primitive feature in speech synthesis. As the context in which a phoneme occurs affects the speech sound, neighboring phonemes have conventionally been taken into account. We would like to understand what kinds of phonemes are more remarkable and how the identities of neighboring phonemes affect the representations of current phonemes.
The overall accuracy of classification of previous, current, and next phoneme identities were $73.1\%$, $84.0\%$, and $67.1\%$, respectively. This suggests that a previous phoneme affects the representation of a current phoneme slightly more than the next one. For more details, Fig.~\ref{bigresults} (a) shows the accuracy per appearance frequency of each phoneme. The general trend is that the classification accuracy clearly drops at each vowel even if the appearance of such phonemes is fairly frequent.
In fact, the prediction accuracy for the encoder outputs derived from consonants was $88.1\%$ on average, while it was $70.7\%$ for those from vowels.
The result phonetically makes sense. In speech, the acoustic quality of vowels is sometimes perceived as weakening because of the physical limitations of the speech organs (e.g., the tongue), which cannot move fast enough to deliver a full-quality vowel.
Vowel reduction is only seen in a spoken language, but the effect appears here in encoded "text" as the drops in the representation quality of vowels. This can be considered as the result of the joint optimization which passes the quality of acoustic features to the encoder outputs.

\subsubsection{Evaluation of syllable features ($b_1$, $b_{16}$)}
English is a stressed-timed language, so stress is a prominent syllable level feature in English TTS systems. Even though lexical stress in English is truly unpredictable and must be memorized along with the pronunciation of an individual word, we found that the trained classifier was able to attain $86.3\%$ accuracy on whether an encoder output was derived from a phoneme in a stressed syllable. The result shows that lexical stress is fairly influential in the encoder outputs, but it is also probably confused with a different level of stress (e.g., prosodic stress), resulting in a reduction in accuracy.
In relation to stress that is caused by the properties of a vowel, it is interesting to see the presence of a vowel at the syllable level. In Fig.~\ref{bigresults} (b), we plot a confusion matrix for classification of vowel identity in the current syllable. The classifier gave a mere $63.8\%$ accuracy on this task. This result is attributed to the same tread of phoneme level contexts where vowels are less prominent than consonants, while the accuracy drops at rarely observed phonemes (i.e., \texttt{AW}, \texttt{OY}, \texttt{UH}) can be ignored.

\subsubsection{Evaluation of POS tagging ($e_1$)}
Part-of-speech (POS) is a commonly used higher feature that associates acoustic modeling with the grammatical structure of a given sentence. In Fig.~\ref{bigresults} (c), we plot a confusion matrix for POS tagging results. While tags for pronouns, determiners, and conjunctions are correctly classified without trouble, much of the misclassification can be seen among nouns, verbs, adjectives, and adverbs. This follows the fact that a lot of words among such parts look alike on the surface. For example, there are denominal adjectives and verbs that are derived from a noun and only differ in their suffix (e.g., wood - wooden). This syntactic and phonemic resemblance causes the encoder outputs of phonemes within such words to be more like each other, making them hard to classify.

\subsubsection{Evaluation of positional features ($b_4$, $p_6$)}
It is important to recognize the position of each symbol to read at multiple levels because a rise or fall in speech quality due to pitch often occurs at linguistic and phonetic boundaries (boundary tone). Fig.~\ref{bigresults} (d) compares prediction accuracy of the phoneme positions in a syllable with the syllable positions in a word.  About the higher accuracy of the syllable positions at the end of words than at the beginning of words, a probable explanation for this is speech quality changes frequently at the end of a sentence (i.e., a group of words), such as in interrogative sentences, and this makes encoder outputs in the syllables near the end of words more distinctive than others.  Also, we found a reduction in accuracy at the middle of the phoneme positions in a syllable. This is possibly because vowels that are less distinctive in representations are likely to be located near the middle of a syllable (nucleus).

\section{Conclusions}
\label{sec:conclusions}
In this work, we investigated how and what types of context information are used in an end-to-end TTS system by comparing its feature representations with the contexts used in parametric TTS. Our experiments revealed the contexts that play an important role in parametric TTS were also remarkable in encoder outputs of end-to-end TTS. Furthermore, it turned out that encoder outputs embrace more detailed information about various levels of context features. The main factors of such effects are the joint optimization of context and acoustic features as well as the generative model that captures long-term structure.
This work provides a unique viewpoint to understand state-of-the-art speech synthesis. The insights gained in this work will be helpful to develop new strategies for the augmentation of an encoder, conditioning it more effectively on various contexts.
\clearpage
\vfill\pagebreak

\bibliographystyle{IEEEbib}
\bibliography{refs}
\end{document}